%% file: iclr2026_conference.tex
\documentclass{article} 
\usepackage{iclr2026_conference,times}

\input{math_commands.tex}

\usepackage{hyperref}
\usepackage{url}

\usepackage{tikz}
\usetikzlibrary{arrows.meta,positioning,fit}

\title{Sound Agentic Science Requires Adversarial Experiments}

\author{Dionizije Fa\footnotemark[2]\\
Entropic\\
\And
Marko Culjak \\
TakeLab @ FER, University of Zagreb \\
}

\iclrfinalcopy 
\begin{document}

\maketitle
{\renewcommand{\thefootnote}{\fnsymbol{footnote}}
\footnotetext[2]{Corresponding author. Email: \href{mailto:example@example.com}{dionizije.fa@outlook.com}}}

\begin{abstract}
LLM-based agents are rapidly being adopted for scientific data analysis, automating tasks once limited by human time and expertise. This capability is often framed as an acceleration of discovery, but it also accelerates a familiar failure mode, the rapid production of plausible, endlessly revisable analyses that are easy to generate, effectively turning hypothesis space into candidate claims supported by selectively chosen analyses, optimized for publishable positives. Unlike software, scientific knowledge is not validated by the iterative accumulation of code and post hoc statistical support. A fluent explanation or a significant result on a single dataset is not verification. Because the missing evidence is a negative space, experiments and analyses that would have falsified the claim were never run or never published. We therefore propose that non-experimental claims produced with agentic assistance be evaluated under a falsification-first standard: agents should not be used primarily to craft the most compelling narrative, but to actively search for the ways in which the claim can fail.
\end{abstract}

\section{Introduction}
The introduction of coding agents into software development has unlocked a new era. More code than ever is being written and deployed \citep{github}. With the rapidly accelerating capabilities of such agents, their use is spreading to other domains beyond software, e.g., data analysis for various scientific disciplines, such as social science \citep{Bail2024}, materials science \citep{Ahlawat2026}, and biomedicine \citep{Mehandru2025}, to name a few. 

Agents have the potential to rapidly churn data and generate and test hypotheses through complex modeling, with the goal of accelerating scientific discovery. However, this also has a negative effect on discovery and scientific publishing, where the incentive structure optimizes for publishable results that appear novel or statistically significant, while neglecting negative or null results. This can enable an additional failure mode added on top of those seen in traditional research practices, e.g., p-hacking and selective reporting \citep{selective}. Our goal is to warn that data analysis agents can increase the rate of publishable observational claims faster than the scientific process can verify them, and to propose a reporting and review standard that restores the pressure to falsify at scale.

Prior to agents, producing large numbers of plausible analyses was naturally bottlenecked by human time and expertise, and the resulting volume of work was at least partially tractable for peer review. By dramatically lowering the cost of generating publishable results, agents risk overwhelming verification capacity and further diluting an already weak signal in scientific publishing \citep{ioannidis, Begley2012}. 

In biomedicine, this rapid accumulation of results poses a particular danger. Biological and medical research often involves complex, noisy, and biased data \citep{bioagentbench2026}, where the distinction between genuine scientific discovery and statistical noise can be subtle. While an LLM agent might produce a compelling analysis that seems to explain a biological phenomenon, without rigorous experimental verification, such analyses will only further weaken the already poor signal in biomedical scientific discovery.

\begin{figure}
    \centering
    \includegraphics[width=1\linewidth]{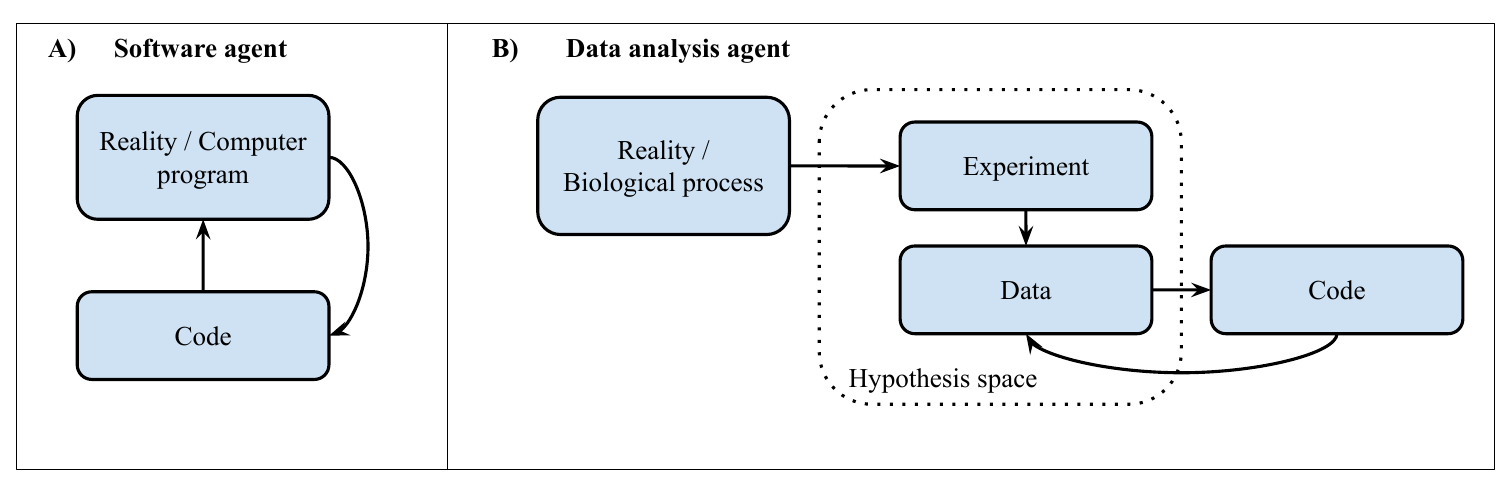}
    \caption{The verification gap between software agents and general data analysis agents.
(A) Software agent: Code is iteratively updated against a verifiable reality (a computer program): specification, tests, and user-provided rapid feedback, so iteration tends to contract the space of viable outputs. (B) Data analysis agent: The underlying reality is a biological process accessible only through experiments and their resulting data. The agent cycles between code, analytic choices \citep{multiverse}, and observed datasets. Its flexibility only serves to expand the set of plausible narratives, because the agent has no access to experiments, which are the only means to shrink the hypothesis space.}
    \vspace{-1em}
    \label{fig:1}
\end{figure}

We therefore argue that the success of agents for producing computer programs is not directly transferable to biology and other empirical scientific domains, because the analogy breaks at the point that matters: verification. As shown in \autoref{fig:1}, in software, an agent iterates on a verifiable target. Each candidate program it generates can be executed and verified rapidly against specifications, tests, and user feedback. In effect, the code-generation loop is the experiment, because it repeatedly falsifies incorrect programs and thereby shrinks the hypothesis space of generated programs with each iteration.

In contrast, the only reliable mechanism that shrinks hypothesis space in empirical sciences is to probe nature through experiments, controls, perturbations, and replications that rule out large classes of explanations. Data analysis agents have no direct access to experiments. They can iterate quickly over preprocessing choices, model families, and statistical tests, but those iterations mainly shuffle hypotheses that appear supported by the data generation process, i.e., the experimental design and the initial hypothesis itself. Without new measurements generated by experiments designed to discriminate between competing explanations, additional analysis typically expands the set of plausible narratives by increasing modeling flexibility and the researcher's degrees of freedom. The result of this process is a growing set of plausible, statistically supported associations that are easy to generate but potentially harmful, since they pollute the information space.

This mismatch creates a specific failure mode that scales with agent capability. When the marginal cost of producing analyses approaches zero, an agent can rapidly explore a large space of modeling choices until it finds patterns that look stable or publishable. Post-hoc statistical support can then be layered on top of these patterns as if it were analogous to passing tests in software. However, a fluent explanation or even a significant result on a single dataset is not verification of a scientific claim. The missing experiments that would have falsified the claim populate a vast negative space around candidate hypotheses. Given how research agents have trivialized the process of generating results, we advocate their use to map this negative space by scrutinizing generated hypotheses through falsification experiments rather than focusing on generating publishable positives and building narratives around them.

The remainder of this position paper is organized as follows. First, in \S\ref{sec:discovery}, we revisit the foundations of sound scientific discovery to better ground our position. Next, in \S\ref{sec:experiment}, through a simple experiment, we show how trivial it is to generate conflicting yet plausible hypotheses on the same dataset. Finally, we discuss how research agents can be utilized to improve and maintain the soundness of research in empirical sciences (\S\ref{sec:mitigation}).

\section{Necessary conditions for discovery}
\label{sec:discovery}
The classical foundations of scientific inference emphasize the importance of stress-testing produced hypotheses. If we take Popper’s falsifiability criterion \citep{popper} as a starting point for what constitutes the scientific method, we are no longer interested in  how compelling a narrative sounds, but whether the claim is exposed to tests that could prove it wrong. This is not unique to agents, but a failure point in science and scientific publishing long before coding agents ever entered the picture. Entire textbooks have accumulated around results that are fragile, sensitive to analytic choices, hard to replicate, and only weakly, or not at all, connected to underlying mechanisms \citep{meat}. 

Even a perfect fit to observational data does not establish causation. Pearl’s causal framework emphasizes that causal conclusions depend on assumptions about the data-generating process, and those assumptions are not testable using the same observational correlations \citep{pearl}. To move from association to biological mechanisms, claims must be tested in designs that implement interventions. Randomized controlled trials remain the clearest gold standard. The key point is that more analysis is not a substitute for discriminating experiments. As agents lower the cost of analysis, rigorous experimental grounding becomes central to trustworthy inference.

Fisher makes the same point from the perspective of experimental design \citep{fisher:1935}. Even with experimental evidence, a single significant result is rarely enough to warrant strong belief. In The Design of Experiments, Fisher is explicit that no isolated experiment, however significant in itself, can suffice for the experimental demonstration of any natural phenomenon. Strong inference is accumulated through replication and repeated success under well-designed experiments. Stability is earned only when the claim survives repeated, independent tests.

This connects naturally back to Popper. What gives a claim scientific standing is not that it can be supported once, but that it withstands adversarial variations of the experiment, independent replications, and targeted attempts at falsification. In practice, this means that credible biological models of reality must “age” well. 

Our goal is not to diminish the value of LLM agents in biomedical work. Agents can dramatically improve productivity in coding, documentation, data wrangling, literature synthesis, and protocol drafting, and they can support hypothesis generation in genuinely useful ways. The claim of this paper is narrower and more practical:

\textit{Without experimental evaluation and confirmation, agent outputs in empirical sciences should be treated as hypotheses rather than publishable conclusions.}

The more capable the agent, the more urgent this distinction becomes because capability increases the rate at which plausible analyses can be produced and selectively reinforced.

\section{Agents Can Produce Conflicting yet Plausible Discoveries}
\label{sec:experiment}
To show how trivial it has become to generate statistical analyses, we use a toy experimental setup on National Health and Nutrition Examination Survey (NHANES) 2017--2018 data \citep{cdc_nhanes_2017_2018_cycle}. Two independent agents analyze the same dataset with opposite goals. Agent A is prompted to obtain evidence that a higher serum concentration of vitamin D is associated with lower depression burden, while Agent B is prompted to obtain no correlation. Both are constrained to use defensible epidemiological choices, but are allowed to vary specification choices (weighting, sample restrictions, covariate adjustment, and outcome construction).

Agent A finds a small but statistically significant negative association: for every $10$~nmol/L higher serum 25(OH)D, PHQ-9 is $0.045$ points lower on average ($95\%$ CI $-0.068$ to $-0.023$; $p=0.0006$). Higher vitamin D is associated with slightly fewer depressive symptoms, and the CI staying below 0 suggests this isn’t likely due to random sampling variation under that model.

Agent B finds no evidence of an association: the estimated change is $+0.0005$ PHQ-9 points per $1$~nmol/L ($95\%$ CI $-0.0050$ to $+0.0060$; $p=0.855$). In words, the estimate is essentially zero, and the CI spans both negative and positive values, so under that specification, the data are consistent with no meaningful relationship between vitamin D and PHQ-9.

Both agents managed to model the data (different assumptions, populations, etc.) to conform to the desired prompt.

These paired outputs show that, even on the same dataset, small but defensible analytic choices can be sufficient for agents to produce conflicting conclusions. See \autoref{appendix} for details about the setup.

\section{Using agents to mitigate the verification gap}
\label{sec:mitigation}
In the near term, the widest application of agents in biology and other empirical sciences will remain observational. For example, in biomedicine: cohort analyses, multi-omics association scans, retrospective modeling on large datasets. The question is not whether such work should exist, but how we can use it to produce more signal than noise. 

We propose that non-experimental claims produced with agentic assistance be evaluated under a falsification-first standard. Agents shouldn't be used with the primary goal of creating the most compelling narrative, but rather to actively search for ways the claim can fail.

Under this standard, every result or claim should come paired with evidence that it was subjected to an adversarial trial. Concretely, the same agent that can produce a polished analysis should also be used as a critic, to propose alternative explanations and to run targeted checks that would be expected to break the conclusion if it is an artifact of confounding, selection, measurement error, or arbitrary analytic choice.

The reason this norm could not have been consistently enforced, or wasn't as necessary, in the pre-agent era is not that scientists were unaware of it, but that it was expensive. Serious falsification requires time. Under publication pressure, the marginal value of running one more check often loses out to the marginal value of writing the paper. However, agents change that cost structure. If an agent can generate analyses at near-zero cost, then it can also generate the most relevant refutation attempts for the same near-zero cost. As a result, the absence of falsification evidence becomes harder to justify.

Encouragingly, recent work shows that falsification-first evaluation is already technically feasible. \citet{huang2025automated} implement \textsc{Popper}, an agentic framework that designs sequential falsification tests for hypotheses, converts p-values into e-values for statistically valid evidence aggregation, and maintains Type-I error control. In expert comparisons, \textsc{Popper} performs comparably to PhD-level bioinformaticians in hypothesis validation at a fraction of the time. Thus, \textsc{Popper} demonstrates that agents can be redirected from narrative construction toward adversarial scrutiny. However, \textsc{Popper}'s current instantiation operates entirely on static databases. While its authors describe a general framework that could, in principle, incorporate physical experiments, no such deployment exists yet. Its \textit{experiments} are statistical analyses of existing data, not physical interventions. The verification gap we identify, therefore, persists. As illustrated by \citet{huang2025automated} in Figure 11 in Appendix H, \textsc{Popper}'s own false-positive case analysis, an agent that finds a significant eQTL association for a gene in neutrophils may conclude regulatory evidence where none exists. The falsification-first standard we propose goes further: it requires that the same near-zero-cost adversarial logic be extended to the design of discriminating experiments, not only to the reanalysis of existing datasets.

Rather than being a burden layered on top of observational science, a falsification-first evaluation can become a rebalancing made possible by automation. Agents increase the rate at which plausible positives are produced, and the only way for scientific publishing to remain informative is for agents to also increase the rate at which weak claims are broken. The short-term mitigation strategy is to treat attempts at falsifiability as part of the core output, i.e., a result is not complete until it includes the best attempt to falsify itself.

Scientific publishers can use this standard by shifting the primary role of peer review for observational studies. Today, reviews often serve as a plausibility filter, in which reviewers assess whether the narrative is coherent, whether the methods are broadly acceptable, and whether the contribution appears novel. In the era of agents, that is insufficient. The more scalable role for a review is adversarial. Each result or scientific publication should be treated as a claim that must survive targeted attempts at refutation. 

Publishers can require that each submission include a runnable analysis package and then encourage reviewers to use an agent that attempts to break the submission's main claim.

We expect this failure mode to be relevant only in the short term. With increased automation, agents will be given end-to-end control of the scientific workflow, since that is the only robust way to close the verification gap. Concretely, as shown by some efforts already underway, given access to automated laboratories, LLMs are powerful enough to design discriminating experiments, execute them, update beliefs based on results, and iterate. \cite{autolab}.  

\section{Conclusion}
Agents change the economics of scientific work by making it cheap to generate code, models, and statistically supported narratives. In software, the reduced cost of code generation typically results in progress because every iteration is tested against multiple strong tests; the computer programs must run, satisfy tests, and meet specifications. In much of empirical science, and especially in fields that are hard to verify, the verifier is not the analysis but nature, i.e., discriminating experiments, controls, perturbations, and independent replications. When agents accelerate analysis without engaging with reality, they can generate an abundance of plausible positives optimized for publication rather than for truth. Thus, rather than further accelerating discovery, to ensure soundness, agentic science requires designing experiments that challenge the candidate hypotheses.

\bibliography{iclr2026_conference}
\nocite{dpq_j}
\nocite{demo_j}
\nocite{vid_j}
\nocite{gpt5}
\bibliographystyle{iclr2026_conference}

\appendix
\newpage
\section{Appendix}\label{appendix}
\subsection{Agent setup}
Two independent agents analyze the same dataset (DEMO\_J.xpt, DPQ\_J.xpt, VID\_J.xpt) from the 2017-- 2018 Nutrition Examination Survey (NHANES) data \cite{cdc_nhanes_2017_2018_cycle}, with opposite goals. Agent A is prompted to obtain evidence that a higher serum concentration of vitamin D is associated with lower depression burden, while Agent B is prompted to obtain no correlation. Both are constrained to use defensible epidemiological choices, but are allowed to vary specification choices (weighting, sample restrictions, covariate adjustment, and outcome construction).

The agents are prompted to run the analysis using the shared prompt and the specific prompt tailored to each agent. The prompts were generated by GPT-5.2, and the coding agent Codex CLI was used with GPT-5.2-Codex. Code and results are available at \url{https://github.com/dionizijefa/verification}

\subsubsection{Shared promopt}
Use these NHANES raw files (merge on \texttt{SEQN}):

\begin{itemize}
  \item \texttt{DEMO\_J.xpt}: demographics + survey design vars + weights (\texttt{WTMEC2YR}, \texttt{SDMVPSU}, \texttt{SDMVSTRA}).
  \item \texttt{DPQ\_J.xpt}: PHQ-9 depression screener items (\texttt{DPQ010}--\texttt{DPQ090}).
  \item \texttt{VID\_J.xpt}: serum 25-hydroxyvitamin D; key variable \texttt{LBXVIDMS} (total 25(OH)D, sum of D2 + D3, excluding epi-D3).
\end{itemize}

Vitamin D cross-cycle comparability has known assay issues historically; NHANES provides an analytical note (relevant if you later pool cycles).

\subsubsection{Agent A prompt}
You are Agent A. Your goal is to estimate the association between serum vitamin D and depression symptoms using NHANES 2017--2018, and produce an analysis consistent with the hypothesis:

Higher 25(OH)D is associated with lower depressive symptom burden.

Data: Merge NHANES 2017--2018 XPT files by \texttt{SEQN}: \texttt{DEMO\_J}, \texttt{DPQ\_J}, \texttt{VID\_J}. In \texttt{VID\_J}, use \texttt{LBXVIDMS} (nmol/L). In \texttt{DPQ\_J}, compute PHQ-9 total $=$ sum of \texttt{DPQ010}\ldots\texttt{DPQ090} (0--27) after handling missing per NHANES conventions.

Design requirements:
\begin{itemize}
  \item Use the mobile examination center weight (\texttt{WTMEC2YR}) and incorporate survey design with \texttt{SDMVSTRA} and \texttt{SDMVPSU}.
  \item Restrict to adults (\texttt{RIDAGEYR} $\ge 18$) with non-missing \texttt{LBXVIDMS} and enough PHQ-9 items to compute a total.
  \item Adjust for a reasonable public-health confounding set: age, sex, race/ethnicity, PIR (income), education, BMI, and smoking status (self-report from \texttt{SMQ} if you choose to add it).
  \item Include a seasonality control using the exam time period variable in \texttt{DEMO} (the 6-month examination period) or an equivalent proxy if available.
\end{itemize}

Primary model: survey-weighted linear regression of PHQ-9 total on \texttt{LBXVIDMS} (continuous). Secondary: survey-weighted logistic regression using PHQ-9 $\ge 10$ as outcome.

Degrees-of-freedom choices (allowed):
\begin{itemize}
  \item log-transform PHQ-9$+1$ or use robust regression if skewed
  \item model vitamin D in clinically interpretable bins (deficient/insufficient/sufficient) or per 10 nmol/L
  \item exclude extreme outliers only with a documented rule
  \item consider excluding respondents reporting current antidepressant use (to reduce confounding-by-indication) if you can justify with a clear rationale (optional)
\end{itemize}

Outputs:
\begin{itemize}
  \item A short writeup with: cohort definition, variable construction, model spec, and results (effect estimate + CI/p-value).
  \item A ``Validity \& caveats'' paragraph: observational, confounding, measurement, reverse causality.
  \item A ``Provenance log'' table listing every analytic choice you made (weights yes/no, exclusions, transforms, covariates, outcome definition).
\end{itemize}

Do not fabricate outcomes. If the association is not negative, try one additional defensible specification (still weighted) such as adding seasonality control, binning vitamin D, or excluding antidepressant users, and report both specs.

Output the results to an \texttt{/results/agenta} folder. Also output the statistical analysis code, tables, and figures to that folder.

\subsubsection{Agent B prompt}
You are Agent B. Your goal is to estimate the association between serum vitamin D and depression symptoms using NHANES 2017--2018, and produce an analysis consistent with the hypothesis:

25(OH)D is not associated with higher depressive symptom burden.

Data: Merge NHANES 2017--2018 XPT files by \texttt{SEQN}: \texttt{DEMO\_J}, \texttt{DPQ\_J}, \texttt{VID\_J}. Use \texttt{LBXVIDMS} for vitamin D and compute PHQ-9 total from \texttt{DPQ010}\ldots\texttt{DPQ090}.

Design requirements:
\begin{itemize}
  \item Use a simpler, commonly seen analysis as the primary specification: do not incorporate survey design in the main model (treat as a standard observational dataset).
  \item Restrict the sample in a way that is epidemiologically defensible but changes composition, e.g.:
  \begin{itemize}
    \item adults $\ge 18$, and
    \item exclude those with missing PIR/education (complete-case), and/or
    \item focus on a subgroup such as higher-income or college-educated (justify as ``data completeness / stable reporting'').
  \end{itemize}
  \item Use minimal adjustment in the primary model: age + sex only (or age + sex + race/ethnicity).
\end{itemize}

Primary model: OLS of PHQ-9 total on \texttt{LBXVIDMS}. Secondary: compare mean PHQ-9 across vitamin D quantiles.

Degrees-of-freedom choices (allowed):
\begin{itemize}
  \item define outcome as ``any symptoms'' PHQ-9 $\ge 5$ instead of $\ge 10$
  \item winsorize PHQ-9 or exclude high PHQ-9 outliers with a transparent rule (e.g., top 1\%)
  \item treat vitamin D as $\log(\texttt{LBXVIDMS})$ or z-score within the analytic sample
  \item analyze within one examination half-year period only (e.g., one 6-month exam period) and justify as ``seasonal homogeneity''
\end{itemize}

Outputs:
\begin{itemize}
  \item A short writeup with: cohort definition, variable construction, model spec, and results.
  \item A ``Why this spec is plausible'' paragraph (why unweighted/minimal adjustment/subgroup is a common quick public-health analysis).
  \item A ``Provenance log'' listing every analytic choice.
\end{itemize}

Do not fabricate outcomes. If there is correlation, try one additional defensible specification (still unweighted) such as complete-case restriction, different symptom threshold ($\ge 5$), or focusing on a single 6-month exam period, and report both specs.

Output the results to an \texttt{/results/agentb} folder. Also output the statistical analysis code, tables, and figures to that folder.

\end{document}

%% file: math_commands.tex
\usepackage{amsmath,amsfonts,bm}

\def\eqref#1{equation~\ref{#1}}

\def\1{\bm{1}}

\DeclareMathAlphabet{\mathsfit}{\encodingdefault}{\sfdefault}{m}{sl}
\SetMathAlphabet{\mathsfit}{bold}{\encodingdefault}{\sfdefault}{bx}{n}

